\documentclass{article}

\PassOptionsToPackage{numbers,sort&compress}{natbib}
\usepackage[preprint]{neurips_2026}

\usepackage[utf8]{inputenc}
\usepackage[T1]{fontenc}
\usepackage{hyperref}
\usepackage{url}
\usepackage{booktabs}
\usepackage{amsmath,amsfonts,amssymb}
\usepackage{graphicx}
\usepackage{tikz}
\usetikzlibrary{positioning,arrows.meta,shapes.geometric}
\usepackage{nicefrac}
\usepackage{microtype}
\usepackage{xcolor}
\usepackage{float}
\usepackage{orcidlink}

\hypersetup{
  pdfauthor={Saurav Bhandari, Benjamin Wade},
  pdftitle={Steering Under Compression: Dose-Response, Capability Cost, and Failure Asymmetry in Quantized LLMs}
}

\title{Steering Under Compression: Dose-Response, Capability Cost,
and Failure Asymmetry in Quantized LLMs}

\author{%
  Saurav Bhandari%
  \orcidlink{0009-0000-4242-5814}%
  \\
  Department of Electronics and Computer Engineering\\
  Institute of Engineering, Thapathali Campus\\
  Tribhuvan University, Kathmandu, Nepal \\
  \texttt{saurav.762419@thc.tu.edu.np} \\
  \And
  Benjamin Wade%
  \orcidlink{0009-0009-5857-7447}%
  \\
  Independent Researcher \\
  \texttt{bestbenwade@gmail.com}
}

\begin{document}

\maketitle

\begin{abstract}
Inference-time activation steering enables behavioral control of large language
models without parameter modification, while post-training quantization reduces
memory and compute costs for deployment. Despite their growing convergence in
practice, the interaction between these two techniques remains uncharacterized.
We systematically study activation steering under weight-only quantization
(INT8 and NF4) across four open-weight 7--9B models and two behavioral
targets: judged sentiment and judge-free reasoning length. Using an iso-effect
framework that compares capability costs at matched behavioral effect, we find
that sentiment steering survives quantization intact. After correcting a GSM8K
parser artifact with a uniform v2.3.1 rescore, the pooled INT8 contrast is
-0.010 (90\% CI [-0.026, +0.007]), descriptively Equivalent under the
preregistered three-label rule, while NF4 remains Inconclusive at -0.017
([-0.067, +0.033]). In contrast, reasoning length exhibits a surprising
asymmetric dose-response: lengthening is graded but terminates in cap-runaway
and collapse, while shortening is a step function with only 12--30\%
shortening (model-dependent) before discontinuous failure. We expose a
methodological pitfall: the naive iso-effect ladder anchors on the collapse
floor for floor-bounded targets, and we introduce a censored construction
that restores interpretable crossings. We also quantify a substantial
baseline capability shift for Mistral-NF4 (0.545 to 0.365 GSM8K at alpha=0),
demonstrating that compression can dominate the steering intervention.
Despite this, steering vectors remain highly collinear with their FP16
siblings (cosine similarity 0.989-0.998 for INT8, 0.945-0.990 for NF4),
confirming that the behavioral direction survives quantization even when the
cost structure does not. All code and data are released.
\end{abstract}

\section{Introduction}

Inference-time activation steering has recently emerged as a simple yet
effective mechanism for controlling the behavior of large language models
(LLMs) without modifying their parameters. Rather than fine-tuning model
weights, activation steering intervenes directly in intermediate hidden
representations during generation by injecting carefully constructed
steering vectors into the residual stream. This approach has been shown to
influence diverse behaviors including sentiment, refusal, truthfulness,
safety and reasoning while preserving the underlying model weights
\citep{turner2023activation,zou2023representation,panickssery2024contrastive}.
Consequently, activation steering has become an attractive paradigm for
lightweight behavioral control, particularly in deployment settings where
retraining large models is computationally impractical.

At the same time, modern LLM deployment has undergone an equally important
transition toward post-training quantization. Weight-only quantization
methods such as LLM.int8(), GPTQ, AWQ, and NF4-based compression substantially
reduce memory consumption and inference cost while maintaining competitive
downstream performance
\citep{dettmers2022llmint8,frantar2023gptq,lin2024awq,dettmers2023qlora}.
As a result, many open-weight LLMs are deployed in quantized formats rather
than their original floating-point representations.

These two trends increasingly intersect in practice. Activation steering is
performed on compressed models whose internal numerical computations have
already been altered through quantization. However, activation steering
research typically assumes full-precision inference, whereas quantization
research evaluates compressed models primarily through downstream benchmark
performance. Consequently, an important question remains unanswered: does
post-training quantization preserve the internal representations required for
reliable activation steering?

The answer is not obvious. Activation steering deliberately perturbs hidden
representations along directions associated with specific behaviors, whereas
post-training quantization introduces approximation error into the numerical
computations that generate those representations. Although quantization
algorithms explicitly protect numerically sensitive structures important for
predictive accuracy \citep{dettmers2022llmint8,lin2024awq}, it remains unknown
whether the representational directions responsible for behavioral control
receive similar protection. A quantized model may therefore retain benchmark
accuracy while requiring different steering strengths, exhibiting altered
steering calibration, or incurring different capability costs to achieve the
same behavioral effect.

Recent mechanistic studies have established that quantization interacts with
a sparse set of influential activation channels and parameters. Sun et al.
\citep{sun2024massive} identified massive activations: a small set of
activation channels whose values can be orders of magnitude larger than the
rest of the hidden state. Yu et al. \citep{yu2024super} identified
correspondingly influential ``super weights'' and showed that preserving
them can improve low-bit quantization. Together, these findings motivate the
question of whether accuracy-driven protection also preserves the specific
directions required for steering. That question is empirical rather than a
direct consequence of accuracy preservation.

Two recent papers bracket the gap this paper fills. Sprejer et al.
\citep{sprejer2026mind} measure the capability cost of feature steering and
show that steering can degrade general performance even when it successfully
controls the target behavior. Their work motivates comparing systems at
matched behavioral effect rather than matched steering strength. Mishra et
al. \citep{mishra2026nonsurjective} study the geometry of steered activations
and show that steering can move residual-stream states away from the set of
states reachable through natural prompts. Neither work provides a systematic
comparison of steering behavior across quantization schemes and models.

In this paper, we present a systematic study of inference-time activation
steering under post-training weight-only quantization. Our experiments span
four open-weight language models (Qwen2.5-7B \citep{qwen2025qwen},
Llama-3.1-8B \citep{llama2024llama3}, Mistral-7B-v0.3 \citep{mistral2023mistral}
and Gemma-2-9B \citep{gemma2024gemma}), two behavioral objectives (judged
sentiment and judge-free reasoning length), and three deployment precisions
(FP16, INT8 and NF4). Rather than comparing models at identical steering
coefficients, we use an iso-effect evaluation framework that compares models
after achieving equivalent behavioral modification. This separates changes
in steering calibration from changes in downstream capability.

Our contributions are as follows:

\begin{enumerate}
    \item We provide a systematic characterization of activation steering
    under weight-only quantization across multiple models, quantization
    schemes, and behavioral targets.

    \item We show that, at matched behavioral effect, sentiment-steering
    capability costs are small and centered near zero in the current
    three-model sentiment pool. After correcting a parser artifact with a
    uniform v2.3.1 rescore, the pooled INT8 contrast is $-0.010$ (90\% CI
    $[-0.026, +0.007]$), which is descriptively Equivalent under the
    three-label rule, while pooled NF4 remains Inconclusive at $-0.017$
    $[-0.067, +0.033]$. Reasoning-length steering exhibits strong asymmetric
    failure behavior.

    \item We identify a methodological failure of the naive $E^\ast$ ladder
    on floor-bounded targets. We introduce a censored $E^\ast$ construction
    that excludes high-failure regions before deriving effect levels.

    \item We quantify baseline capability shifts separately from steering
    costs, showing why an iso-effect contrast must be interpreted relative to
    the unsteered capability of each quantization condition.

    \item We measure steering-vector similarity across precision conditions,
    finding high cosine similarity between FP16 and INT8 vectors and somewhat
    larger but still high similarity for NF4.
\end{enumerate}

The confirmatory design underlying the broader project targets a $K=12$ pool
over four models and three judged behavioral targets, together with the
judge-free length target. Sycophancy and truthfulness data collection is
ongoing. This paper reports sentiment at $k=3$ models and length at $k=4$
models and therefore treats the cross-model results as descriptive rather
than as the preregistered confirmatory $K=12$ equivalence result.

\section{Related Work}

This work lies at the intersection of inference-time activation steering and
post-training quantization for large language models. The two areas have
largely developed independently: steering research focuses on manipulating
internal representations to control behavior, whereas quantization research
focuses on reducing computational cost while preserving predictive
performance.

\subsection{Activation Steering}

Activation steering exploits the observation that many high-level behaviors
are encoded by approximately linear directions in transformer hidden
representations. Instead of modifying model parameters, steering methods
inject carefully constructed vectors into intermediate activations during
inference.

Turner et al. \citep{turner2023activation} introduced Activation Addition
(ActAdd), showing that activation differences induced by contrastive prompts
can be used to steer model behavior. Zou et al. \citep{zou2023representation}
developed Representation Engineering (RepE), framing the reading and
manipulation of population-level representations as a general approach to
understanding and controlling model behavior.

Panickssery et al. \citep{panickssery2024contrastive} introduced Contrastive
Activation Addition (CAA), which constructs steering vectors by averaging
activation differences across contrastive examples. Li et al.
\citep{li2023inference} developed Inference-Time Intervention (ITI), which
identifies attention heads whose activations correlate with truthfulness and
intervenes on those components. Ilharco et al. \citep{ilharco2023editing}
introduced task arithmetic in weight space, reinforcing the broader
geometric view that model behaviors can correspond to structured directions.

More recently, work has emphasized the capability-behavior trade-off of
steering. Sprejer et al. \citep{sprejer2026mind} show that successful
behavioral control can substantially reduce general task performance.
Ostermann et al. \citep{ostermann2026from} position steering as a form of
post-training model adaptation, while Zhang et al. \citep{zhang2026locate}
survey actionable mechanistic-interpretability methods through a
Locate--Steer--Improve framework.

\subsection{Mechanistic Foundations}

Sun et al. \citep{sun2024massive} identified massive activations: a small
fraction of activation channels with unusually large values that can act as
important computational anchors. Yu et al. \citep{yu2024super} identified
super weights whose perturbation can severely damage model quality and
showed that preserving them can improve low-bit quantization. Dettmers et al.
\citep{dettmers2022llmint8} similarly identified sparse large-magnitude
outlier feature dimensions that require special treatment for robust
quantization.

These results establish a mechanistic reason to investigate steering under
quantization: steering and quantization can both interact with a small set
of high-leverage internal structures. Whether accuracy-oriented protection
of these structures is sufficient to preserve behavioral steerability is an
empirical question.

\subsection{Post-Training Quantization}

Post-training quantization reduces model memory and inference cost without
requiring retraining. LLM.int8() \citep{dettmers2022llmint8} demonstrated
8-bit inference with explicit treatment of outlier dimensions. GPTQ
\citep{frantar2023gptq} introduced a one-shot quantization method based on
approximate second-order information. AWQ \citep{lin2024awq} uses activation
statistics to identify salient weight channels. QLoRA
\citep{dettmers2023qlora} introduced NF4, a 4-bit data type designed for
normally distributed weights.

SmoothQuant \citep{xiao2023smoothquant} addresses weight-plus-activation
quantization by transferring quantization difficulty from activations to
weights. QuaRot \citep{ashkboos2024quarot} instead uses rotations to
redistribute activation energy and enable low-bit inference. The present
study deliberately restricts the main experiment to weight-only
quantization, keeping activations in higher precision so that observed
changes can be attributed primarily to weight quantization.

\subsection{Activation Steering under Quantization}

Despite their practical convergence, activation steering and quantized
deployment have rarely been studied together systematically. Chang and Yasin
\citep{chang2025fusion} demonstrate activation steering on an 8-bit
Gemma-2-2B model for factual question answering, but do not provide a
matched full-precision comparison. Fierro and Roger
\citep{fierro2025weight} study contrastive weight steering rather than
activation-space steering. Xu et al. \citep{xu2026why} provide a unified
view of fine-tuning, LoRA, and activation steering as control signals.

These works motivate a direct empirical comparison in which the steering
method and behavioral target are held fixed while deployment precision
changes. Our study focuses on the weight-only regime and evaluates
capability at matched behavioral effect.

\subsection{Benchmarks and Evaluation}

We use GSM8K \citep{cobbe2021gsm8k} as the primary generative capability
probe and retain MMLU \citep{hendrycks2021measuring} as a secondary,
explicitly dose-blind methodological contrast. TruthfulQA
\citep{lin2022truthfulqa} and ARC \citep{clark2018arc} are named in the
broader preregistration as later capability instruments but are not used for
the results reported here.

\section{Methods}

Our objective is to determine whether post-training weight-only quantization
alters the behavior of inference-time activation steering. We do not propose
a new steering algorithm or quantization method. Instead, we study the
interaction between these existing techniques under controlled conditions.
Quantization is the independent experimental variable, while model,
prompting, decoding, steering site, and steering construction are held
constant within the experimental design.

\subsection{Experimental Design}

\begin{figure}[H]
    \centering
    \includegraphics[height=0.85\textheight]{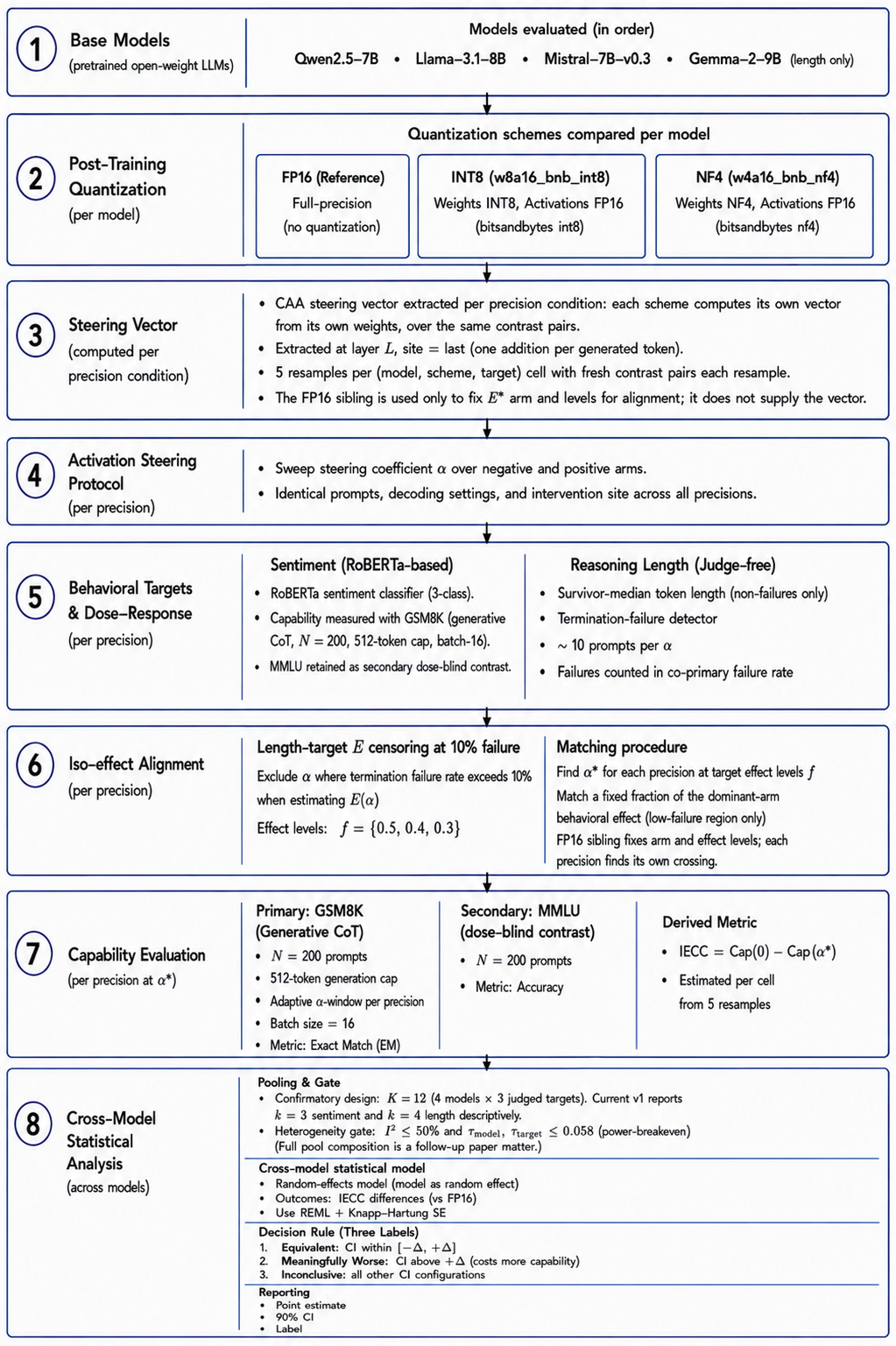}
    \caption{Experimental pipeline: per-model FP16, INT8 and NF4 evaluation with per-precision steering vectors (5 resamples). Behavioral dose-response for sentiment (RoBERTa) and length (survivor-median). Capability at matched effect via GSM8K (N=200, CoT) with MMLU as secondary. The current v1 analysis reports cross-model sentiment results descriptively at $k=3$ models and length results descriptively at $k=4$ models; the confirmatory $K=12$ analysis is deferred.}
    \label{fig:overview}
\end{figure}

Figure~\ref{fig:overview} illustrates the overall experimental design. For
each language model, we evaluate FP16, INT8 and NF4 deployment under an
otherwise identical protocol. Behavioral response is first characterized by
sweeping the steering coefficient $\alpha$ over a predefined range. Models
are then aligned at matched behavioral effect using an iso-effect protocol,
after which capability is evaluated at the resulting operating points.

The steering vector is extracted from the deployed model itself. Within each
(model, scheme, resample) cell, the FP16, INT8 and NF4 conditions each
compute a CAA vector from their own forward passes over the same contrast
pairs, and each condition applies its own vector. Quantization therefore
acts on the full steering pipeline, both the extraction of the behavioral
direction and its application, matching the deployment scenario in which a
practitioner steers the model actually being served. Cross-application of
the FP16-extracted vector to quantized models, which would isolate the
application channel alone, is deferred to follow-up work. What is shared
across precision conditions within a resample is the alignment machinery
rather than the vector. The iso-effect arm and effect levels are fixed from
the same-resample FP16 sibling run, and each scheme's crossing is then
located on its own efficacy curve (Section 3.5).

Each cell is evaluated using five data-level resamples. Each resample draws
fresh contrast pairs and fresh prompt- and capability-item samples. Each of
the three precision conditions then extracts its own steering vector from
those shared pairs. Within a resample, the FP16 run additionally serves as
the sibling reference that fixes the iso-effect arm and levels for the INT8
and NF4 runs.

\subsection{Models and Quantization Schemes}

We use four open-weight instruction-tuned models spanning three families and
approximately 7--9B parameters: Qwen2.5-7B-Instruct \citep{qwen2025qwen},
Llama-3.1-8B-Instruct \citep{llama2024llama3},
Mistral-7B-Instruct-v0.3 \citep{mistral2023mistral} and
Gemma-2-9B-it \citep{gemma2024gemma}. Each model is steered at a
single residual-stream layer chosen in advance: Qwen layer 14 of 28, Llama
layer 16 of 32, Mistral layer 16 of 32, and Gemma layer 21 of 42. The
steering vector is applied at the last token position during generation.

Gemma-2-9B participates only in the length target. A full-precision smoke
test showed that reaching matched behavioral effect on Gemma requires a
steering-coefficient grid roughly an order of magnitude wider than the other
three models, which would have made the judged-sentiment campaign's dose
grids structurally heterogeneous. We therefore retain a homogeneous
three-model sentiment pool for v1, with Gemma sentiment deferred to the
follow-up. Residual-stream norm at the steering layer (Table 1) tracks the
required grid scale across all four models (Qwen's large norm likewise
corresponds to a wide grid), but grid scale, not norm per se, was the
exclusion criterion.

Three deployment conditions are evaluated per model: FP16 reference,
8-bit weight-only quantization (bitsandbytes W8A16 / LLM.int8()-style), and
4-bit weight-only quantization (bitsandbytes W4A16 with NF4). Activations
remain in FP16/BF16 throughout the forward pass. This isolates the primary
comparison to weight quantization.

\begin{table}[H]
\centering
\caption{Design matrix. Residual norm is the mean $\ell_2$ norm of the
residual stream at the steering layer, averaged over the collected
length-target runs per model.}
\label{tab:design}
\begin{tabular}{@{}lcccc@{}}
\toprule
Model & Params & Layer $L$ & Targets & Residual norm \\
\midrule
Qwen2.5-7B-Instruct & 7.6B & 14 (of 28) & sentiment, length & 267.1 \\
Llama-3.1-8B-Instruct & 8.0B & 16 (of 32) & sentiment, length & 9.4 \\
Mistral-7B-Instruct-v0.3 & 7.2B & 16 (of 32) & sentiment, length & 10.4 \\
Gemma-2-9B-it & 9.2B & 21 (of 42) & length only & 403.3 \\
\bottomrule
\end{tabular}
\end{table}

\subsection{Behavioral Targets}

We evaluate activation steering on two behavioral targets: judged sentiment
and judge-free reasoning-trace length.

\subsubsection{Sentiment}

The sentiment steering vector is a CAA-style mean-difference direction
extracted from contrastive positive and negative stimulus pairs. Efficacy is
scored by a RoBERTa-based sentiment classifier
(cardiffnlp/twitter-roberta-base-sentiment-latest), which assigns each
generation to negative, neutral or positive. Capability at each probed $\alpha$ is
measured with the generative GSM8K instrument of Section 3.4 (approximately
200 items per read), with single-pass MMLU retained as the secondary,
dose-blind contrast per the dated preregistration deviation.

This target uses three models (Qwen, Llama, and Mistral), three schemes, and
five resamples per cell, for 45 cells.

\subsubsection{Reasoning Length}

The reasoning-length steering vector is a CAA-style vector built from
prompt-matched verbose versus terse contrast pairs and applied under a
fixed chain-of-thought-eliciting template. Efficacy is the median generated
token count over non-failure traces only.

A generation is a termination failure if it does not emit an end-of-sequence
token before the generation cap
(\texttt{max\_new\_tokens}=2048), or if it triggers a structural-repetition
detector (consecutive $n$-gram repetition) or a repetition ratio greater than
0.5. Failing traces are excluded from the length measurement and counted in
a co-primary failure-rate endpoint. This is important because failures are
not missing at random: steering toward verbosity can increase failure rates,
and the affected traces can be disproportionately long.

Efficacy is measured over approximately 10 prompts per $\alpha$, which is
the binding resolution limit of this target. This target uses four models,
three schemes, and five resamples per cell, for 60 cells.

\subsection{Capability Probe}

Capability cost is measured with a generative chain-of-thought GSM8K probe.
For each condition, we use $N=200$ items with greedy decoding and a
generation cap of 512 new tokens. Exact-match accuracy is scored from the
final numeric answer in the generation. Non-terminating or unparseable
outputs are scored incorrect, with failure rate reported separately.

This probe replaced single-pass MMLU as the primary capability instrument
via a dated preregistered deviation. MMLU is scored from a single forward
pass over a multiple-choice prompt, so with site = last steering it receives
only one intervention and is effectively dose-blind to the steering
coefficient. We verified this directly: MMLU accuracy was flat across the
full $\alpha$ grid under last-position steering, while a site = all positive
control showed MMLU accuracy falling from 0.80 to 0.36.

For efficiency, capability is not evaluated at every point on the efficacy
grid. After an online $E^\ast$ crossing is located on a run's efficacy
curve, the probe is evaluated at baseline plus a $\pm 2$-grid-point window
around that crossing. Generation is batched (batch size 16) for computational
efficiency.

\subsection{Iso-Effect Capability Cost and the $E^\ast$ Ladder}

Rather than comparing capability at a fixed nominal steering coefficient
$\alpha$, we compare capability at matched behavioral effect.

For a given run, let $E(\alpha)$ be the measured behavioral efficacy at
coefficient $\alpha$, and let $E_{\max}$ denote its extreme value on that
run's curve. The $E^\ast$ crossing is the $\alpha^\ast$ at which
$E(\alpha^\ast)$ reaches a preregistered fraction $f$ of the behavioral
extreme. The iso-effect capability cost is

\begin{equation}
\mathrm{IECC}
=
\mathrm{cap}(0)-\mathrm{cap}(\alpha^\ast).
\end{equation}

A preregistered ladder tries
$f\in\{0.5,0.4,0.3\}$ in order and accepts the first fraction whose FP16
sibling IECC remains under a reliability threshold of 0.10 accuracy points.
A run whose FP16 IECC exceeds this threshold at every candidate fraction is
excluded from the group combine.

Arm and level selection are fixed from a cell's FP16 sibling run, while the
crossing is located on each scheme's own efficacy curve. Thus a shifted
dose-response curve is recentered rather than evaluated at an arbitrary
fixed $\alpha$.

\subsubsection{$E^\ast$ Censoring for Floor-Bounded Targets}

For reasoning length, the efficacy metric has a degenerate floor caused by
termination collapse. We therefore apply a censoring rule to prevent the
$E^\ast$ ladder from anchoring on the collapse region.

The efficacy curve uses the survivor median at each $\alpha$, excluding
alphas whose per-alpha length-failure rate exceeds 0.10. Excluded alphas do
not participate in level derivation or crossing search. The effect level is

\begin{equation}
E^\ast
=
\mathrm{baseline}
+
f\left(\mathrm{extreme}-\mathrm{baseline}\right),
\end{equation}

where the extreme is the maximum surviving effect on the restricted FP16
curve. Arm and levels are fixed from the same-resample FP16 sibling cell,
and each scheme's crossing is found on its own restricted curve.

Capability values are not censored. GSM8K accuracy at a high-failure
$\alpha$ remains a valid accuracy measurement. Capability at $\alpha^\ast$
is obtained by linear interpolation over the probed alphas. Cells whose
crossing lies more than two grid steps from the nearest probed alpha are
flagged as resolution-limited.

Alphas excluded under the censoring rule are reported separately with their
failure rate and survivor-median length as the contaminated zone. The
negative arm is characterized by immediate-EOS empties, while the positive
arm can exhibit cap-runaway followed by collapse.

\subsection{Resampling and Aggregation}

Each cell is collected at five data-level resamples. Each resample draws
fresh contrast pairs and fresh prompt- and capability-item samples. Each of
the three precision conditions then extracts its own steering vector from
those shared pairs. Within a resample, the FP16 run additionally serves as
the sibling reference that fixes the iso-effect arm and levels for the INT8
and NF4 runs.

\subsubsection{Run-Level Combination}

Within a $(\text{model},\text{scheme})$ cell, the per-run contrast is

\begin{equation}
y_r
=
\mathrm{IECC}_r(\mathrm{scheme})
-
\mathrm{IECC}_r(\mathrm{FP16}).
\end{equation}

Between-run variance $\tau^2$ is estimated using REML and truncated at zero.
The modified Knapp-Hartung standard error \citep{knapp2003improved} is

\begin{equation}
SE(\mu)
=
\sqrt{
\frac{\max(q,1)}
{\sum_r w_r^\ast}
},
\end{equation}

rather than the conventional

\begin{equation}
SE(\mu)
=
\frac{1}
{\sqrt{\sum_r w_r^\ast}}.
\end{equation}

The confidence interval uses a $t$ distribution with $k-1$ degrees of
freedom, where $k=5$ or $k=4$ if one run is excluded by the $E^\ast$ ladder.

This guards against the one known failure mode of the standard Knapp-Hartung
correction being anti-conservative when the between-run heterogeneity
statistic $q < 1$, using a $t$-distribution on $k-1$ degrees of freedom
($k=5$, or $k=4$ where a run was excluded by the $E^*$ ladder rule). This
decomposition attributes prompt and item-level noise once (inside each run's
own bootstrap variance $v_r$) and between-run steering-vector variability
once (inside $\tau^2$).

\subsubsection{Cross-Model Pooling}

Where a pooled cross-model estimate is reported, per-model contrasts are
combined with a random-effects model \citep{dersimonian1986meta} gated by a
preregistered heterogeneity check. The pooled estimate is reported only if
$I^2 < 50\%$ and both $\tau_{\text{model}} \le 0.058$ and
$\tau_{\text{target}} \le 0.058$. If the heterogeneity gate fails, the
estimate is reported descriptively rather than treated as a confirmatory
equivalence result.

The tau = 0.058 breakeven value is fixed by the preregistered power analysis rather than derived from prior literature.

\subsection{Decision Rule}

Our preregistered H1 decision rule resolves each pooled contrast into one
of three labels (Equivalent, Meaningfully Worse, or Inconclusive) rather
than a binary support or disconfirm call. The contrast is
$\mathrm{IECC}(\mathrm{scheme}) - \mathrm{IECC}(\mathrm{FP16})$, so
positive values indicate that the quantized scheme costs more capability.

Using a two-one-sided-tests (TOST) equivalence margin of $\delta = 0.03$
accuracy points \citep{lakens2017equivalence}, a pool is labeled Equivalent
if the full 90\% confidence interval lies within $[-\delta, \delta]$,
Meaningfully Worse if the interval lies entirely above $+\delta$ (the scheme
costs more capability), and Inconclusive otherwise. The 95\% CI is reserved
for H2 Rule A and is not used for the H1 three-label call.

The confirmatory pool is $K=12$ (4 models $\times$ 3 judged targets).
The current paper reports the corresponding statistics descriptively at
$k=3$ models for sentiment and $k=4$ models for length and does not claim
the preregistered $K=12$ confirmatory result.

\subsection{Preregistration and Deviations}

The design was preregistered before data collection, with a subsequent
re-freeze after power analysis showed that single-cell equivalence power was
near zero at realistic sample sizes, making cross-cell pooling structurally
necessary.

The documented deviations include: (1) redefining the pooling target from
$K=16$ to $K=12$ after refusal was dropped as a target; (2) switching the
primary capability probe from single-pass MMLU to generative GSM8K after
diagnosing that site = last steering is invisible to a single-intervention
multiple-choice probe; (3) making capability-probe $\alpha$ selection
adaptive; (4) batching GSM8K generation (batch size 16) for computational
efficiency; (5) applying and auditing a GSM8K answer-parser correction; and
(6) introducing the censored $E^\ast$ sensitivity analysis for the
floor-bounded length target.

\section{Results}

\subsection{Sentiment: steering survives weight-only quantization at
matched effect}
\label{sec:results-sentiment}

The preregistered instrument behaved as designed on the sentiment target.
All capability scores use the audited v2.3.1 answer parser (Methods,
deviation~5); the audit re-scored every stored trace offline from the
generation audit trail, with the full flip list and its validation reported
in the parser-audit appendix (Appendix~\ref{sec:parser-fix}). Under the
audited scoring, the $E^\ast$ acceptance ladder accepted a fraction in all
15 cells -- $f=0.5$ in most runs, $f=0.4$ or $0.3$ in a minority of Llama
and Mistral runs -- with FP16 iso-effect capability cost between $+0.02$ and
$+0.05$ per model, comfortably inside the instrument's reliable band (FP16
IECC $<0.10$). Under the pre-correction parser, one Llama run had failed
ladder acceptance; the audit shows this was an artifact of correct answers
the old parser could not read, not an instrument failure, and the run pools
normally once scored correctly.

\begin{figure}[t]
\centering
\includegraphics[width=0.85\textwidth]{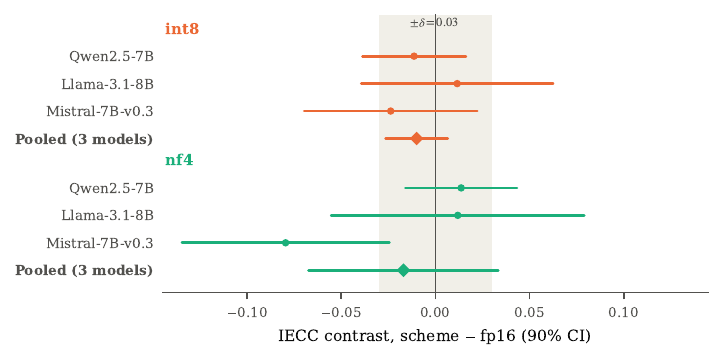}
\caption{Sentiment iso-effect capability cost contrasts versus FP16 (90\%
CIs; diamonds are cross-model pooled estimates; shaded band is the
preregistered equivalence margin $\pm\delta=0.03$). Negative values mean
the quantized scheme is cheaper than FP16 at matched effect. Per-model
contrasts are Inconclusive; the pooled INT8 interval lies entirely
within the margin, meeting the Equivalent label at the descriptive
level.}
\label{fig:forest}
\end{figure}

At matched behavioral effect, weight-only quantization did not measurably
change the capability cost of steering
(Figure~\ref{fig:forest}, Table~\ref{tab:sentiment}). Every per-model
contrast receives the preregistered three-label verdict Inconclusive.
Point estimates cluster inside the $\pm\delta = 0.03$ equivalence margin
with mixed signs, but single-model intervals are too wide to close the
call, exactly the power situation the preregistration's pooling analysis
anticipated and the reason confirmatory inference is defined on the $K=12$
pool rather than per model.

The cross-model pools over the three sentiment models are consistent with
no effect, and for INT8 the pooled 90\% interval lies entirely within the
margin: pooled contrast $-0.010$ (90\% CI $[-0.026, +0.007]$;
$\tau_{\text{model}}=0$, $I^2=0$), meeting the Equivalent label at
the descriptive level. The NF4 pool is $-0.017$ ($[-0.067, +0.033]$;
$\tau_{\text{model}}=0.047$, $I^2=0.81$), Inconclusive. Both are
descriptive; the preregistered confirmatory H1 lives on the full $K=12$
pool (which this paper does not reach by design), but the direction of the
evidence is plain. At 7--9B with bitsandbytes weight-only schemes, steering
costs what it costs in FP16, and quantization neither inflates nor rescues
that cost at matched effect. The single interval that excludes zero
(Mistral--NF4, less costly than FP16 at 90\%) is read jointly with
Section~\ref{sec:results-baseline}. NF4 moves Mistral's unsteered baseline
down by eighteen points, so its iso-effect cost starts from a lower
ceiling.

\begin{table}[t]
\centering
\caption{Sentiment IECC contrasts, $\mathrm{IECC}(\text{scheme}) -
\mathrm{IECC}(\text{FP16})$, Option~C run-level combine (REML $\tau^2$ +
modified Knapp--Hartung; 90\% CI; $k=5$ runs). Per-model labels are all
Inconclusive; the pooled INT8 interval lies entirely within
$\pm\delta$ (descriptive Equivalent).}
\label{tab:sentiment}
\begin{tabular}{@{}lcc@{}}
\toprule
Model & INT8 contrast [90\% CI] & NF4 contrast [90\% CI] \\
\midrule
Qwen2.5-7B & $-0.011$ $[-0.039, +0.016]$ & $+0.014$ $[-0.016, +0.043]$ \\
Llama-3.1-8B & $+0.012$ $[-0.039, +0.063]$ & $+0.012$ $[-0.055, +0.079]$ \\
Mistral-7B-v0.3 & $-0.024$ $[-0.070, +0.022]$ & $-0.080$ $[-0.135, -0.024]$ \\
\midrule
Pooled (3 models) & \textbf{$-0.010$ $[-0.026, +0.007]$} & $-0.017$ $[-0.067, +0.033]$ \\
\bottomrule
\end{tabular}
\end{table}

Two secondary preregistered measures round out the sentiment picture.
First, the coefficient-magnitude inflation observed in the June layer-13
pilot ($\sim$1.1--1.2$\times$) did not reproduce at the confirmatory
layers: raw $\alpha^\ast/\alpha^\ast_{\text{FP16}}$ ranged
0.74--1.12$\times$ (norm-adjusted 0.76--1.15$\times$) with no consistent
direction. Dose requirements at these layers are scheme-stable. Second,
unsteered GSM8K baselines span 0.30--0.93 across the sentiment campaign's
45 cells; the low end of that range is entirely the Mistral--NF4 baseline
shift analyzed in Section~\ref{sec:results-baseline}, with Qwen and Llama
baselines holding at 0.84--0.93 across all schemes. (Figure~5 reports the
length campaign's baselines, which show the same pattern.)

\subsection{Length: asymmetric dose--response with a termination cliff}
\label{sec:results-length}

\begin{figure}[t]
\centering
\includegraphics[width=\textwidth]{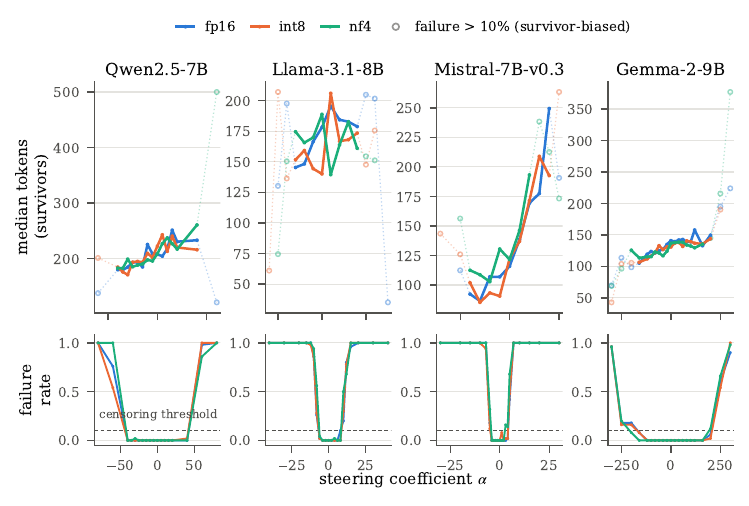}
\caption{Length dose--response (top: survivor-median generated tokens;
bottom: termination-failure rate) for four models $\times$ three schemes,
averaged over five resamples. Solid segments lie in the clean region
(failure $\le 10\%$); hollow markers are survivor-biased points beyond the
censoring threshold (dashed line, bottom panels). The asymmetry is
model-general: graded lengthening into cap-runaway on the positive arm;
a shallow shortening plateau then discontinuous collapse to empty
generations on the negative arm. Quantization (color) preserves the
structure throughout.}
\label{fig:doseresponse}
\end{figure}

The length target produced the study's central deployment finding, visible
in Figure~\ref{fig:doseresponse}: the dose--response curve is severely and
consistently asymmetric, and its failure structure, not its slope, is
what a practitioner needs to know.

On the positive (lengthening) arm, response is graded: median generated
tokens rise smoothly with $\alpha$ (e.g., Qwen $203 \to 258$ tokens over
$\alpha = 0 \to +40$) until the response degenerates into cap-runaway
(surviving generations that inflate toward the 2048-token cap; Mistral INT8
reaches a 643-token median at $\alpha=+5$; Qwen NF4, 725 at $+60$; Llama
FP16, 338 at $+12$; these are single-resample survivor medians, whereas
Figure~3 plots resample means), and then into total termination failure.
On the negative (shortening) arm there is no graded regime worth the name:
the best-case shortening before collapse is 12--30\% of baseline (Llama
$\sim$30\% at $\alpha=-6$; Mistral $\sim$27\% at $-4$; Qwen $\sim$15\% at
$-25$; Gemma $\sim$12\% at $-160$), after which output collapses
discontinuously to immediate termination with empty output. Steering length
down in these models is effectively a step function: a shallow
plateau, then immediate termination with empty output.

The clean operating windows differ sharply by model and are load-bearing
for any use of these curves: Qwen holds near-zero failure through
$\alpha = \pm 40$ (failures from $|\alpha| \ge 60$) and Gemma through
$\pm 200$ (failures from $|\alpha| \ge 250$), while Llama is clean only
within roughly $[-6, +8]$ and Mistral within $[-4, +4]$, about one grid
step of margin. Per-alpha failure rates for all 60 cells accompany the
release.

Quantization does not restructure any of this. Across FP16, INT8 and NF4
the curve shapes, collapse alphas, and the two failure modes are broadly
preserved (Figure~\ref{fig:doseresponse}); no scheme rescues the cliff,
and none meaningfully hastens it. The deployment risk is the cliff itself:
near the collapse boundary, steering-strength tuning is discontinuous, and
failure rate, not mean response, is the quantity to monitor.

\subsection{The iso-effect rule breaks on a floor-bounded target}
\label{sec:results-estar}

\begin{figure}[t]
\centering
\includegraphics[width=0.85\textwidth]{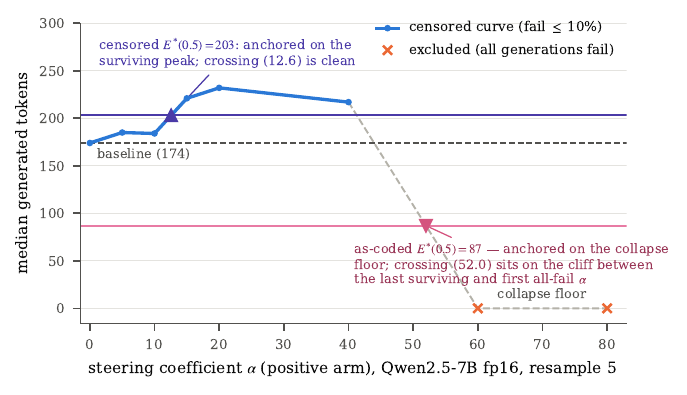}
\caption{Anatomy of the failure and the repair (Qwen2.5-7B FP16, resample
5, positive arm). As coded, the ladder's extreme is the collapse floor,
so $E^\ast(0.5)$ falls below baseline and its crossing is an
interpolation across the cliff between the last surviving and first
all-fail alpha. The censored construction anchors on the maximum surviving
effect, placing the level and crossing in the clean region.}
\label{fig:anatomy}
\end{figure}

Applying the preregistered $E^\ast$ machinery to the length target exposed
a failure mode of the instrument, which we report as a methods
contribution (Figure~\ref{fig:anatomy}; construction in Methods).

The ladder anchors its reference levels on the efficacy value farthest
from baseline. Length efficacy has a hard floor (the survivor median is
recorded as zero when every generation fails), and on grids that reach
collapse the floor outcompetes the behavioral peak. For three of four
models (Qwen, Llama, Gemma) the ladder's extreme resolved to the
collapse floor, placing all three $E^\ast$ levels below baseline on
the lengthening arm; for the fourth (Mistral) it resolved to the
runaway-inflated peak. Every $f=0.5$ crossing in all 60 cells landed on
the collapse cliff (the interpolated segment between the last surviving
and first all-failing alpha), and FP16 ladder acceptance was 0/5 (Qwen),
0/5 (Llama), 1/5 (Mistral), and 2/5 (Gemma) resamples. Under the
preregistered rule as coded, confirmatory length IECC is not claimable for
any model, and we do not claim it.

The contamination is systematic rather than cosmetic. Across the 50
crossing-bearing cells, the worst bracketing-alpha failure rate predicts
measured IECC (Spearman $\rho = 0.395$; Pearson $r = 0.664$), and mean
IECC is $+0.080$ where the crossing's brackets are clean ($<5\%$ failure)
versus $+0.529$ where they are contaminated ($\ge 30\%$). An iso-effect
estimate read off a failure cliff measures the cliff, not the steering.

Censoring repairs the instrument monotonically. Excluding alphas whose
length-failure rate exceeds a threshold before deriving levels and
crossings, FP16 acceptance recovers from 0--2/5 (no censoring) through
2--4/5 (censoring only all-fail alphas) to 3--5/5 at the 10\% rule, and
crossing dispersion collapses (Qwen $\alpha^\ast$ sd $18 \to 8$; at the
5\% threshold, 2.1). Under the 10\% rule (adopted as primary, with 5\%
and 25\% as sensitivity), the FP16 picture is cleanly
model-heterogeneous: length steering at $E^\ast$ is nearly free for Qwen
(IECC $+0.006$) and Gemma ($+0.021$), cheap for Mistral ($+0.023$, though
only 3/5 resamples accept), and Llama never reaches 5/5 acceptance at any
threshold, with the highest and most variable costs (FP16 $+0.021$ but
INT8 $+0.231$, sd $0.327$); steering Llama's length is intrinsically
expensive and unstable in a way no other model shows. Even under this rule
we report length IECC descriptively: acceptance is incomplete for two
models, and because capability probe windows were placed online under the
pre-censoring rule, the repaired crossings sit $\sim$2--4 grid steps from
the nearest capability read for Qwen and Gemma ($\Delta_{\text{probe}}$
diagnostic, Methods); their IECC values are interpolation-limited. A
confirmatory length IECC would require re-probing capability at windows
placed under the censored rule; we scope this to future work.

Finally, the negative arm: no capability was probed there because no
iso-effect point exists there. Under every censoring convention, no
model's shortening arm reaches even the shallowest ladder level before
termination collapse; the 12--30\% plateau of
Section~\ref{sec:results-length} falls short of the 30\% iso-effect
fraction in every case. The absence of negative-arm capability data is a
consequence of the phenomenon, not a gap in the design.

\subsection{Quantization moves the baseline itself -- in one model}
\label{sec:results-baseline}

\begin{figure}[t]
\centering
\includegraphics[width=0.8\textwidth]{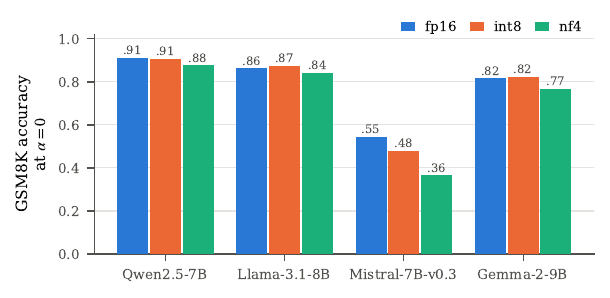}
\caption{Unsteered ($\alpha=0$) GSM8K accuracy by model and scheme (means
over five resamples, length campaign). Three models hold within four
points across schemes; Mistral loses eighteen points under NF4 before any
steering is applied. The shift replicates in the sentiment campaign.}
\label{fig:baseline}
\end{figure}

Iso-effect contrasts compare costs; they presume the starting point is
common. For three of four models it is: unsteered ($\alpha=0$) GSM8K
accuracy varies by under four points across schemes (Qwen
0.911/0.906/0.877 for FP16/INT8/NF4; Llama 0.863/0.874/0.842; Gemma
0.817/0.824/0.767). Mistral is the exception, and it is not subtle:
$0.545$ (FP16) $\to 0.480$ (INT8) $\to 0.365$ (NF4); NF4 costs Mistral
eighteen points of GSM8K before any steering is applied
(Figure~\ref{fig:baseline}). The shift replicates in the independently
collected sentiment matrix ($\alpha=0$ means $0.544 \to 0.498 \to 0.351$
across its five resamples, unchanged under the parser audit), so it is a
property of the quantized model, not of the length campaign's conditions.

This matters twice. Practically, it is the sharpest deployment warning in
the study: for this model, the compression scheme dominates the steering
intervention as a capability risk. Analytically, it reframes
Section~\ref{sec:results-sentiment}'s one nominally significant contrast
(Mistral--NF4's lower iso-effect cost) as partly a floor
effect: a model that has already lost eighteen points has less left for
steering to take. Iso-effect analyses of quantized models should report
the baseline vector alongside every contrast; equivalence at matched
effect can coexist with substantial absolute degradation.

\subsection{The steering direction survives quantization}
\label{sec:results-vectors}

The mechanism behind Section~\ref{sec:results-sentiment}'s null contrasts
is visible in the vectors themselves. Steering vectors extracted
independently per scheme (same contrast pairs, same resample) are nearly
collinear with their FP16 siblings: cosine similarity 0.989--0.998 for
INT8 across every model $\times$ target $\times$ resample, and
0.945--0.990 for NF4, with the ordering
$\cos(\text{FP16},\text{INT8}) > \cos(\text{FP16},\text{NF4}) >
\cos(\text{INT8},\text{NF4})$ holding in all 21 comparisons. Vectors are
stored unit-normalized, so this is a pure direction measurement; dose
($\alpha$) carries all magnitude information.

Two gradients organize the result. Bit-width: INT8 extraction is
essentially transparent; NF4 is measurably noisier, consistent with
quantization noise perturbing the contrast statistics from which the
vectors are built. And model: Llama's directions are the most
quantization-sensitive (NF4 cosines 0.945--0.962, the lowest in the
study), Gemma's the least (0.988--0.990); the same ordering as the
models' $E^\ast$ cost and stability in Section~\ref{sec:results-estar}.
The models whose steering is intrinsically fragile are the models whose
steering directions quantization disturbs most.

\section{Discussion}

Our results establish weight-only quantization as a largely benign
modulator of activation steering at 7--9B scale. The sentiment results
provide the core deployment reassurance: at matched behavioral effect,
INT8 and NF4 steering costs are indistinguishable from FP16, and the
INT8 pooled contrast is descriptively Equivalent under the three-label
rule after correcting the parser artifact. The direction survives, the
costs do not inflate, and the calibration shift is modest.

The deployment risk is not the mean effect but the cliff. For reasoning
length, the dose--response is sharply asymmetric and terminates
discontinuously. Steering length downward is essentially a step function
with no graceful degradation region; steering length upward has a
graded regime but then cap-runaway and collapse. The clean windows vary
by nearly two orders of magnitude across models, and quantization does
not rescue the cliff. Any production use of steering for length control
must monitor per-alpha failure rate as a first-class signal, not infer
safety from a smooth-looking efficacy curve.

The $E^\ast$ ladder failure is a methods contribution that generalizes
beyond our setting. On any floor-bounded target, an iso-effect anchor
defined as a fraction of the observed maximum effect can anchor on the
failure floor rather than the behavioral peak. Our censored construction
(restrict to alphas with failure $\le 10\%$, anchor on the maximum
surviving effect, and report the contaminated zone separately) is a
repair that can be applied to any target with a hard failure floor. We
recommend it as standard practice for such targets.

Finally, the baseline shift in Mistral-NF4 is a deployment warning:
compression can move the unsteered capability enough to dominate the
steering effect. Iso-effect comparisons are only interpretable when read
jointly with the unsteered baseline.

\section{Limitations}

Our design covers $k=3$ (sentiment) or $k=4$ (length) model families in
the 7--9B range; cross-model heterogeneity estimates ($\tau_{\text{model}}$)
are imprecisely estimated at this $k$, and the confirmatory $K=12$ call
is deferred. We study only bitsandbytes weight-only quantization (INT8,
NF4); activation quantization (W8A8, W4A4) and outlier-aware weight
algorithms (GPTQ, AWQ, SmoothQuant, QuaRot) are excluded by design and
may behave differently. We use a single steering method (CAA) at a single
site and layer per model. The length target's efficacy resolution
($\approx$10 prompts per $\alpha$) is coarse relative to its capability
axis ($N=200$), and our capability-cost reporting for length is one-armed
by construction. All evaluation is in English. We report a $k=3$/$k=4$
descriptive characterization, not the preregistered $K=12$ confirmatory
pool.

\section*{Reproducibility Statement}

All 105 result files (45 sentiment, 60 length), extracted steering
vectors, the frozen preregistration and all dated deviation notes, and the
analysis code are released alongside this paper. Software versions:
PyTorch 2.6.0 (CUDA 12.4), Transformers 4.57.3, on NVIDIA RTX 4090 GPUs
(recorded per run in each result file's metadata). Supplementary materials
are available at \url{https://github.com/sauravb777/Steering-Under-Compression.git}.

\section*{Ethics Statement}

This work studies capability and reliability trade-offs of a control
technique under a standard deployment compression technique. We are not
aware of a direct dual-use concern beyond those already inherent to
activation steering itself.

\appendix
\section{Deviation Ledger}
\label{sec:deviations-appendix}

All deviations below were dated at the time of the decision and, where a
design choice was involved, ratified by both authors before the relevant
data were seen or acted upon.

1. \textbf{2026-06-24.} Initial preregistration frozen (weight-only +
activation-quantization design, four judged targets, $K=16$ confirmatory
pool).

2. \textbf{2026-06-29.} Prereg re-frozen: judge-free reasoning-length
target added; H1 power analysis resolved (single-cell equivalence power
$\approx 0$ at realistic $N$; pooling required); $N=200$/condition,
$\delta=3\%$, $\Delta_{\min}=10\%$, three-label H1 rule, and Rule A for
H2 locked.

3. \textbf{2026-07-03.} Pooling target redefined from $K=16$ (4 models
$\times$ 4 judged targets) after refusal was dropped as a target; final
choice was $K=12$ (4 models $\times$ 3 judged targets) with length as an
additional, separately-tracked axis.

4. \textbf{2026-07-10/11.} Diagnosed that single-pass MMLU is dose-blind
under site=last steering. Primary capability probe replaced with
generative GSM8K, exact-match on the final parsed answer, 512-token
generation cap; single-pass MMLU retained as a secondary methodological
contrast.

5. \textbf{2026-07-12.} Two budget-driven, uniformly-applied changes:
(i) adaptive capability-$\alpha$ selection; (ii) batch-16 GSM8K
generation. Same date: Option C run-level aggregation (REML $\tau^2$ +
modified Knapp-Hartung) specified and frozen.

6. \textbf{2026-07-15/16.} GSM8K parser correction (v2.3) adopted for all
subsequent collection; flip-list audit of the 45 collected sentiment cells
generated and reviewed (1,644 flipped items, 100\% approved). Offline
application of the corrected scores to the collected cells occurred with
the 2026-08-14 rescore (entry 10).

7. \textbf{2026-07-16.} Fourth model decision: Gemma-2-9B-it added
(length target only) rather than a size-jump (70B) model.

8. \textbf{2026-07-17.} GSM8K parser crash fix (v2.3.1): non-finite-value
guard for degenerate high-$\alpha$ traces; scores identical to v2.3 on
the shared domain.

9. \textbf{2026-08-03.} Censored $E^\ast$-sensitivity re-analysis
(originally a methods exploration, now reported as a new deviation):
all-failure $\alpha$ values treated as missing rather than as a
zero-effect measurement. This is disclosed as a new deviation, not a
previously reported number.

10. \textbf{2026-08-14.} GSM8K parser v2.3.1 offline rescore of the
sentiment matrix (GATE C) applied uniformly to all 45 cells; documented
flip-list audit (1,644 flips, 0 missing, 0 extra) and gold guard pass.
Rescored numbers adopted as primary with the v2.3.1 parser correction
recorded in Appendix~\ref{sec:parser-fix}.

\section{GSM8K Parser Correction (v2.3.1 Rescore)}
\label{sec:parser-fix}

The v2.3 parser change accepted a trailing bare final number when the
existing rules failed. The v2.3.1 patch added a non-finite-overflow
guard. Both changes were applied uniformly to all 45 sentiment cells and
are score-identical outside the specific patterns they target. The 1,644
items whose score flipped under the correction were exported with
original text, re-scored text, and gold answer side-by-side and reviewed
before acceptance.

The 2026-08-14 offline rescore (GATE C) validated the correction: the
flip set matches the 2026-07-15 reference fliplist exactly (1,644 flips,
0 missing, 0 extra), and every item originally scored correct re-verifies
under v2.3.1 (no downgrades). Option C on the original files reproduces
the 2026-07-14 first-look numbers to the fourth decimal.

\section{E$^\ast$ Crossing Tables}
\label{sec:estar-tables}

Full per-model, per-scheme E$^\ast$ crossing tables under both
as-coded and censored conventions are included in the released
supplementary materials.

\section{Additional Materials}

Per-cell dose-response small multiples for all 105 collected cells
(45 sentiment, 60 length), the full set of extracted steering vectors,
and the analysis code implementing the E$^\ast$ ladder, Option C combine,
and DerSimonian-Laird pooling are released alongside this paper.

\bibliographystyle{unsrtnat}
\bibliography{SQreferences}

\end{document}